\documentclass{article}

\usepackage{todonotes}
\usepackage{times}
\usepackage{graphicx} 
\usepackage{subcaption}
\usepackage{placeins}

\usepackage{natbib}
\usepackage{amsmath}
\usepackage{amssymb}

\usepackage{algorithm}
\usepackage{algorithmic}

\usepackage{hyperref}

\DeclareMathOperator*{\E}{\mathbb{E}}
\DeclareMathOperator{\Tr}{Tr}
\usepackage[accepted]{icml2018}

\icmltitlerunning{Is Generator Conditioning Causally Related to GAN Performance?}

\begin{document} 

\twocolumn[
\icmltitle{Is Generator Conditioning Causally Related to GAN Performance?}




\begin{icmlauthorlist}
\icmlauthor{Augustus Odena}{goo}  
\icmlauthor{Jacob Buckman}{goo}
\icmlauthor{Catherine Olsson}{goo}
\icmlauthor{Tom B. Brown}{goo}
\icmlauthor{Christopher Olah}{goo}
\icmlauthor{Colin Raffel}{goo}
\icmlauthor{Ian Goodfellow}{goo}
\end{icmlauthorlist}

\icmlaffiliation{goo}{Google Brain}

\icmlcorrespondingauthor{Augustus Odena}{augustusodena@google.com}

\icmlkeywords{boring formatting information, machine learning, ICML}

\vskip 0.3in
]



\printAffiliationsAndNotice{} 

\begin{abstract}
Recent work \cite{DYNAMICALISOMETRY} suggests that controlling the entire
distribution of Jacobian
singular values is an important design consideration in deep learning.
Motivated by this, we study the distribution of singular values of the Jacobian
of the generator in Generative Adversarial Networks (GANs).
We find that this Jacobian generally becomes ill-conditioned at the beginning
of training and that the average (with $z \sim p(z)$)
conditioning of the generator is highly predictive of two other ad-hoc metrics
for measuring the ``quality'' of trained GANs: the Inception Score and the
Frechet Inception Distance (FID). We test the hypothesis that this relationship
is causal by proposing a ``regularization'' technique (called Jacobian
Clamping) that softly penalizes the condition number of the generator
Jacobian. Jacobian Clamping improves the mean Inception Score and the mean
FID for GANs trained on several datasets and greatly reduces inter-run
variance of the aforementioned scores,
addressing (at least partially) one of the main criticisms of GANs.

\end{abstract}

\section{Introduction}
\label{section:introduction}

Generative Adversarial Networks (or GANs) are a promising techique for
building flexible generative models \cite{GANS}.
There have been many successful efforts to scale them up to large datasets and
new applications \cite{LAPGAN,DCGAN,ACGAN,STACKGAN,PROGRESSIVEGAN,PROJECTION}.
There have also been many efforts to better understand their training procedure,
and in particular to understand various pathologies that seem to plague
that training procedure
\cite{UNROLLEDGAN,GENERALIZATIONANDEQUILIBRIUM,FID,LOCALLYSTABLE,PRINCIPLED}.
The most notable of these pathologies --- ``mode collapse'' --- is characterized
by a tendency of the generator to output samples from a small subset of the
modes of the data distribution.
In extreme cases, the generator will output only a few unique samples or even
just the same sample repeatedly.
Instead of studying this pathology and others from a probabilistic perspective,
we study the distribution of the squared singular values of the
input-output Jacobian of the generator.
Studying this quantity allows us to characterize GANs in a new
way --- we find that it is predictive of other GAN performance measures.
Moreover, we find that by controlling this quantity, we can improve average-case
performance measures while greatly reducing inter-run variance of those
measures.
More specifically, this work makes the following contributions:

\begin{itemize}

\item We study the squared singular values of the generator Jacobian at
individual points in the latent space.
We find that the Jacobian generally becomes ill-conditioned quickly at the
beginning of training, after which it tends to fall into one of two clusters:
a ``good cluster'' in which the condition number stays the same or even
gradually decreases, and a ``bad cluster'', in which the condition number
continues to grow.

\item We discover a strong correspondence between the conditioning of the
Jacobian and two other quantitative metrics for evaluating GAN quality:
the Inception Score and the Frechet Inception Distance.
GANs with better conditioning tend to perform better according to these metrics.

\item We provide evidence that the above correspondence is causal by
proposing and testing a new regularization technique,
which we call Jacobian Clamping.
We show that you can constrain the conditioning of the Jacobian relatively
cheaply\footnote{The Jacobian Clamping algorithm doubles the batch size.} and
that doing so improves the mean values and reduces inter-run variance of
the values for the Inception Score and FID.

\end{itemize}


\section{Background and Notation}
\label{section:background}


\begin{figure*}[ht]
\includegraphics[width=1.0\textwidth]{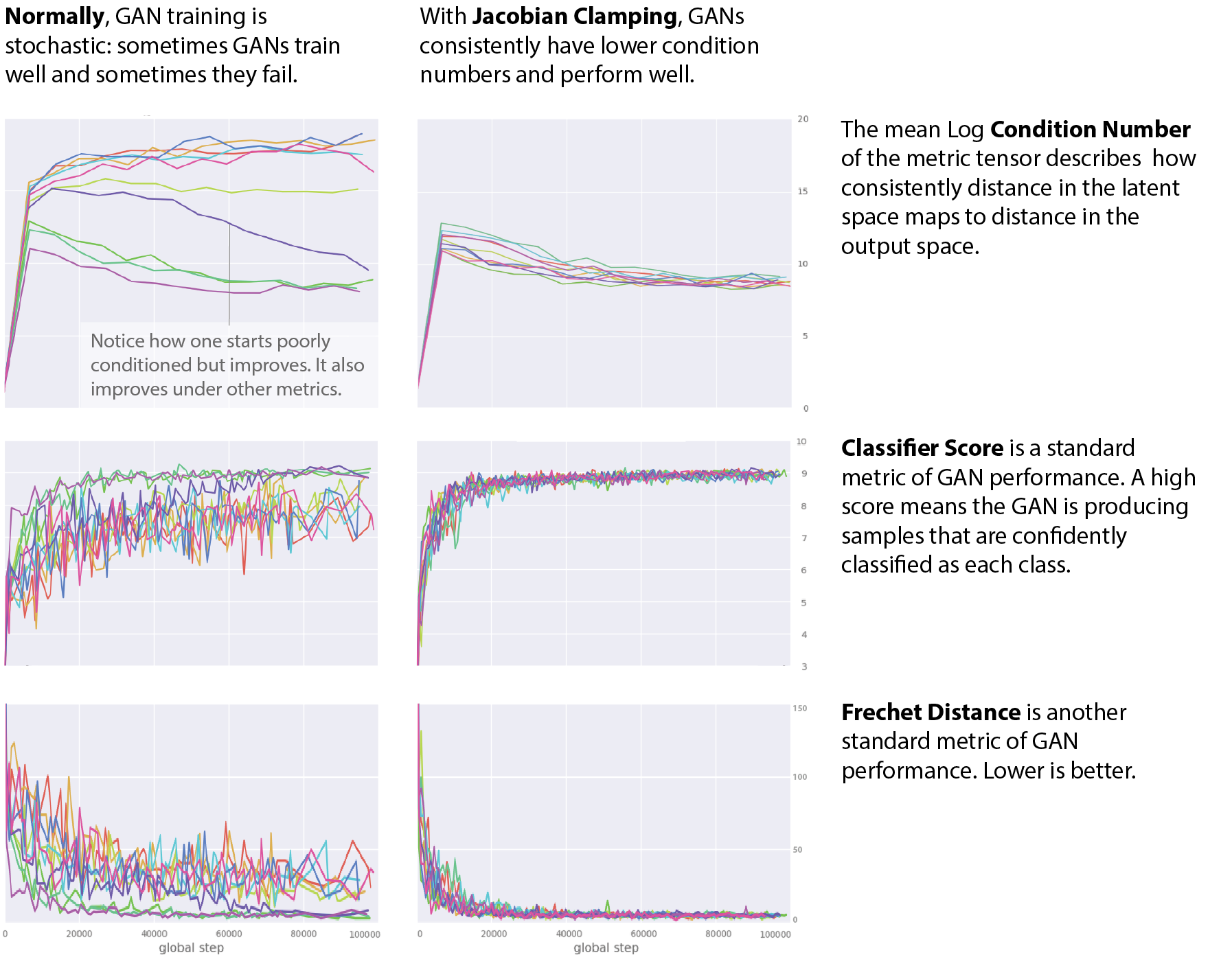}
\caption{
MNIST Experiments. Left and right columns correspond to 10 runs without and with Jacobian Clamping, respectively.
Within each column, each run has a unique color.
Top to bottom, rows correspond to mean log-condition number, Classifier Score, and Frechet Distance. Note the dark purple run in the left column: the
generator moves from the ill-conditioned cluster to the well-conditioned cluster while also moving from the low-scoring cluster to the
high-scoring cluster.
}
\label{fig:mnist}
\end{figure*}

\paragraph{Generative Adversarial Networks:}
A generative adversarial network (GAN) consists of two neural networks trained
in opposition to one another.
The generator $G$ takes as input a random noise vector $z \sim p(z)$ and
outputs a sample $G(z)$.
The discriminator $D$ receives as input either a training sample or a
synthesized sample from the generator and outputs a probability distribution
$D(x)$ over possible sample sources.
The discriminator is then trained to maximize the following cost:
\begin{equation}
L_D = -\E_{x \sim p_\text{data}} [\log D(x)] -
\E_{z \sim p(z)} [\log (1 - D(G(z)))]
\end{equation}
while the generator is trained to minimize\footnote{
This formulation is known as a ``Non-Saturating GAN'' and is the formulation in
wide use, but there are others. See \citet{GANS} for more details.}:
\begin{equation}
L_G = -\E_{z \sim p(z)} [\log D(G(z))]
\end{equation}

\paragraph{Inception Score and Frechet Inception Distance:}
In this work we will refer extensively to two\footnote{
We elect not to use the technique described in \citet{AIS} for reasons explained
in \citet{FLOWGAN}.} ``scores'' that have been proposed
to evaluate the quality of trained GANs. Both make use of a pre-trained
image classifier.
The first is the Inception Score \citep{IMPROVEDTECHNIQUES}, which is given by:

\begin{equation}
\exp\left (\mathbb{E}_{\mathbf{x} \in P_\theta} [KL(p(y \vert \mathbf{x}) \Vert p(y)]  \right)
\end{equation}

where $\mathbf{x}$ is a GAN sample,
$p(y\vert \mathbf{x})$ is the probability for labels $y$ given by a pretrained
classifier on $\mathbf{x}$,
and $p(y)$ is the overall distribution of labels in the generated samples
(according to that classifier).

The second is the Frechet Inception Distance \citep{FID}.
To compute this distance, one assumes that the activations in the coding layer
of the pre-trained classifier come from a multivariate Gaussian.
If the activations on the real data are $N(m, C)$ and the activations on the
fake data are $N(m_w, C_w)$, the FID is given by:

\begin{equation}
\|m-m_w\|_2^2+ \Tr \bigl(C+C_w-2\bigl(CC_w\bigr)^{1/2}\big)
\end{equation}

\paragraph{Mathematical Background and Notation:}
Consider a GAN generator $G$ mapping from latent space with dimension $n_z$
to an observation space with dimension $n_x$.
We can define $Z := R^{n_z}$ and $X := R^{n_x}$ so that we may write
$G : z \in Z \to x \in X$.
At any $z \in Z$, $G$ will have a Jacobian $J_z \in R^{n_x \times n_z}$ where
$(J_z)_{i,j} := \frac{\partial G(z)_i}{\partial z_j}$.
The object we care about will be the distribution of squared singular
values of $J_z$. To see why, note that
the mapping $M: z \in Z \to J_z^TJ_z$ takes any point $z$ to a
symmetric and positive definite matrix of dimension $n_z \times n_z$ and so
constitutes a Riemannian metric.
We will write $J_z^T J_z$ as $M_z$ (and refer to $M_z$ somewhat sloppily as the
``Metric Tensor'').
If we know $M_z$ for all $z \in Z$, we know most of the interesting things
to know about the geometry of the manifold induced by $G$.
In particular, fix some point $z \in Z$ and consider the eigenvalues
$\lambda_1, \dots, \lambda_{n_z}$ and eigenvectors $v_1, \dots, v_{n_z}$ of
$M_z$.
Then for $\epsilon \in R$ and $k \in[1, n_z]$,

\begin{equation}
  \label{eqn:eqlimit}
\lim_{||\epsilon|| \to 0} \frac{||G(z) - G(z + \epsilon v_k)||}{
  ||\epsilon v_k||} = \sqrt{\lambda_k}
\end{equation}

Less formally, the eigenvectors corresponding to the large eigenvalues of
$M_z$ at some point $z \in Z$ give directions in which taking a very
small ``step'' in $Z$ will result in a large change in $G(z)$ (and analogously
with the eigenvectors corresponding to the small eigenvalues).
Because of this, many interesting things can be read out of the eigenspectrum of
$M_z$.

Unfortunately, working with the whole spectrum is unwieldy, so it would be nicer
to work with some summary quantity.
In this work, we choose to study the condition number of $M_z$
(the best justification we can give for this is that we noticed during
exploratory analysis that the condition number was predictive of the Inception
Score, but see the supplementary material for further justification of why we
chose this quantity and not some other quantity).
The condition number is defined for $M_z$ as
$\frac{\lambda_{max}}{\lambda_{min}}$.
If the condition number is high, we say that the metric tensor is
``poorly conditioned''.
If it's low, we say that the metric tensor is ``well conditioned''.

Now note that the eigenvalues of $M_z$ are identical to the squared singular
values of $J_z$. This is why we care about the singular value spectrum of $J_z$.

\section{Analyzing the Local Geometry of GAN Generators}
\label{section:analyzing}

\begin{figure}[ht]
\vskip 0.2in
\begin{center}
\centerline{\includegraphics[width=\columnwidth]{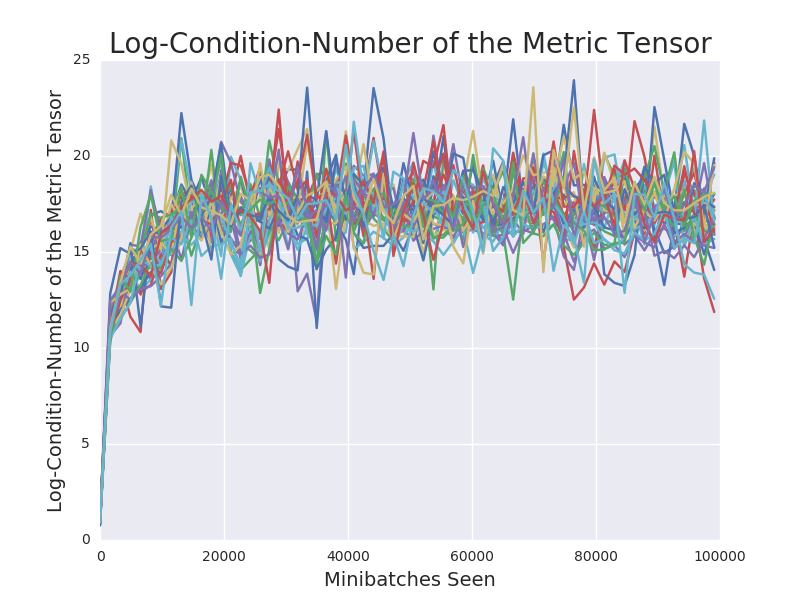}}
\caption{
The condition number of $M_z$ for a GAN trained on MNIST at various fixed $z$
throughout training.
}
\label{fig:condition_number_average}
\end{center}
\vskip -0.2in
\end{figure}

\paragraph{The Metric Tensor Becomes Ill-Conditioned During Training:}
We fix a batch of $z \sim p(z)$ and examine the
condition number of $M_z$ at each of those points as a GAN is training on the
MNIST data-set.
A plot of the results is in Figure \ref{fig:condition_number_average},
where it can be seen that $M_z$ starts off well-conditioned everywhere and
quickly becomes poorly conditioned everywhere.
There is considerable variance in how poor the
conditioning is, with the log-condition-number ranging from around 12
to around 20.

It is natural to ask how consistent this behavior is across different
training runs. To that end, we train 10 GANs that are identical up to random
initialization and compute the average log-condition number across a fixed
batch of $z$ as training progresses (Figure \ref{fig:mnist} Top-Left).
Roughly half of the time, the condition number increases rapidly and then stays
high or climbs higher. The other half of the time, it increases rapidly and
then decreases. This distribution of results is in keeping with the general
understanding that GANs are ``unstable''.

\begin{figure}[ht]
\vskip 0.2in
\begin{center}
\centerline{\includegraphics[width=\columnwidth]{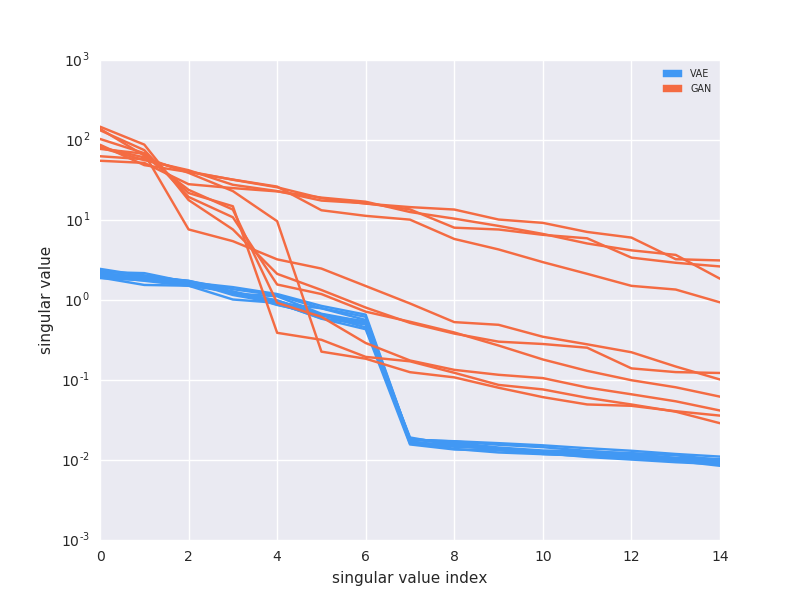}}
\caption{
Log spectra of the average Jacobian from 10 training runs of a variational
autoencoder and 10 training runs of a GAN.
There are a few interesting things about this experiment:
First, it gives a way to quantify how much less `stable' the GAN training
procedure is than the VAE training procedure.
The spectra of the different VAE runs are almost indistinguishable.
Second, though the GAN and VAE decoders both take noise from $N(0, I)$ as input,
the overall sensitivity of the VAE decoder to its input seems to be quite a bit
lower than that of the GAN decoder -- this does not stop the VAE
from successfully modeling the MNIST dataset.
}
\label{fig:vae}
\end{center}
\vskip -0.2in
\end{figure}

\begin{figure*}[ht]
\includegraphics[width=1.0\textwidth]{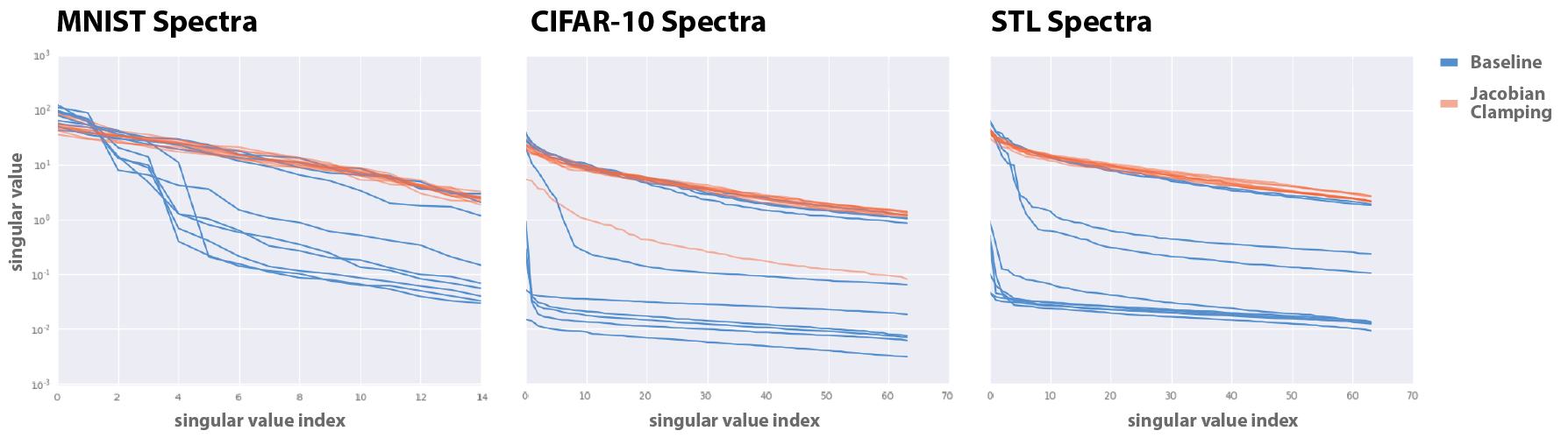}
\caption{
Singular value spectra of the average Jacobian at the end of training, presented
in log-scale.
}
\label{fig:spectra}
\end{figure*}

The condition number is informative and computing its average over many $z$
gives us a single scalar quantity that we can evaluate over time.
However, this is only one of many such quantities that we can compute, and
it obscures certain facts about the singular value spectrum of $J_z$ at various
$z$.
For completeness, we also compute -- following \citet{MCMCVAE},
who does the same for variational autoencoders -- the spectrum
of the average (across a batch of $z$) Jacobian.
It's not clear a priori what one should expect of these spectra,
so to provide context we perform the same computation on 10 training runs of
a variational autoencoder \citep{VAES, VAES2}.
See Figure \ref{fig:vae} for more details.
For convenience, we will largely deal with the condition number
going forward.

\paragraph{Conditioning is Predictive of Other Quality Metrics:}
One reason to be interested in this condition number quantity is that it
corresponds strongly to other metrics used to evaluate GAN quality.

We take two existing metrics for GAN performance and measure how they correspond
to the average log condition number of the metric tensor.
The first measure is the Inception Score \cite{IMPROVEDTECHNIQUES} and the
second measure is the Frechet Inception Distance \cite{FID}.
We test GANs trained on three datasets: MNIST, CIFAR-10, and STL-10
\citep{MNIST,CIFAR,STL}.
On the MNIST dataset, we modify both of these scores
to use a pre-trained MNIST classifier rather than the Inception classifier.
On the CIFAR-10 and STL-10 datasets, we use the scores as defined.
We resized the STL-10 dataset to $48 \times 48$ as has become standard in
the literature about GANs.
The hyperparameters we use are those from \citet{DCGAN}, except that we modified
the generator where appropriate so that the output would be of the right size.

We first discuss results on the MNIST dataset.
The left column of Figure \ref{fig:mnist} corresponds to (the same) 10 runs of
the GAN training procedure with different random initializations.
From top to bottom, the plots show the mean (across the latent space)
log-condition number, the classifier score, and the MNIST Frechet Distance.
The correspondence between condition number and score is quite strong in
both cases.
For both the Classifier Score and the Frechet Distance, the 4 runs with the
lowest condition number also have the 4 best scores.
They also have considerably lower intra-run score variance.
\textbf{Note also that the dark purple run, which transitions over time from
being in the
ill-conditioned cluster to the well-conditioned cluster, also transitions
between clusters in the score plots.}
Examples such as this provide evidence for the significance of the
correspondence.

We conducted the same experiment on the CIFAR-10 and STL-10 datasets.
The results from these experiments can be seen in the left columns of
Figure \ref{fig:cifar} and Figure \ref{fig:stl} respectively.
The correspondence between condition number and the other two scores is also
strong for these datasets. The main difference is that the failure modes on the
larger datasets are more dramatic --- in some runs, the Inception Score never goes
above 1. For both datasets, however, we can see examples of runs with middling
performance according to the score that also have moderate ill-conditioning:
\textbf{In the CIFAR-10 experiments, the light purple run has a score that is
in betweeen the
``good cluster'' and the ``bad cluster'', and it also has a condition number
that is between these clusters.
In the STL-10 experiments, both the red and light purple runs
exhibit this pattern.}

Should we be surprised by this correspondence?
We claim that the answer is yes.
Both the Frechet Inception Distance and the Inception Score are computed
using a pre-trained neural network classifier.
The average condition number is a first-order approximation of
sensitivity (under the Euclidean metric) that makes no reference at all to this
classifier.

\paragraph{Conditioning is Related to Missing Modes:}
\label{section:conditioningisrelated}

Both of the scores aim to measure the extent to which the GAN
is ``missing modes''. The Frechet Inception Distance arguably measures this in a
more principled way than does the Inception Score, but both are designed with
this pathology in mind. We might wonder whether the observed correspondence is
partly due
to a relationship between generator conditioning and the missing-mode-problem.
As a coarse-grained way to test this, we performed the following computation:
Using the same pre-trained MNIST classifier that was used to compute the
scores in Figure \ref{fig:mnist}, we drew 360 samples from each of the 10 models
trained in that figure and examined the distribution over predicted classes.
We then found the class for which each model produced the fewest samples.
The ill-conditioned models often had 0 samples from the least sampled class,
and the well-conditioned models were close to uniformly distributed.
In fact, the correlation coefficient between the mean log condition number for the model
and the number of samples in the model's least sampled class was $-0.86$.

\section{Jacobian Clamping}
\label{section:jacobian}

Given that the conditioning of $J_z$ corresponds to the Inception Score and FID,
it is natural to wonder if there is a causal relationship between these
quantities. The notion of causality is slippery and causal inference is an
active field of research (see \citet{CAUSALITY} and \citet{MAKINGTHINGSHAPPEN}
for overviews from the perspective of computer science and philosophy-of-science
respectively) so we do not expect to be able to give a fully satisfactory answer
to this question.
However, we can perform one relatively popular method for inferring causality
\cite{CAUSALREASONINGTHROUGHINTERVENTION, INTERVENTIONSANDCAUSALINFERENCE},
which is to do an intervention study.
Specifically, we can attempt to control the conditioning directly and
observe what happens to the relevant scores.
In this section we propose a method for accomplishing this control
and demonstrate that it both improves the mean scores and reduces variance of
the scores across runs.
We believe that this result represents an important step toward understanding
the GAN training procedure.

\begin{figure*}[ht]
\includegraphics[width=1.0\textwidth]{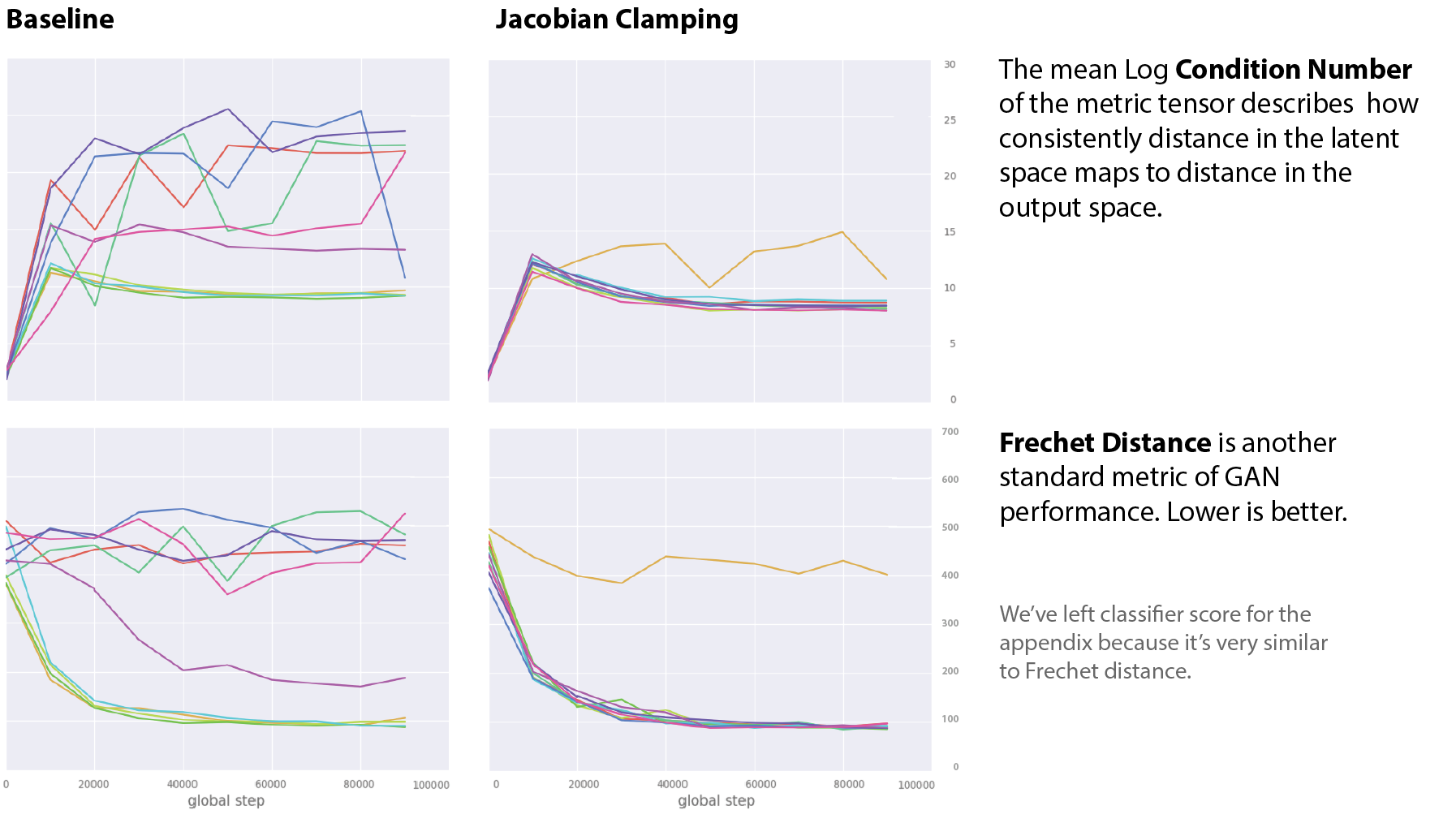}
\caption{
CIFAR10 Experiments. Left and right columns correspond to 10 runs without and with Jacobian Clamping, respectively.
Within each column, each run has a unique color.
Top Row: Mean log-condition number over time. Bottom Row: Frechet Inception Distance over time. Note the light purple run (Left) which has a
condition number between the ill-conditioned cluster and the well-conditioned one; it also has scores between the low-scoring cluster
and the high-scoring one. Note the gold run (Right): it's the only run for which Jacobian Clamping "failed", and it's also the only run for which
the condition number did not decrease after its initial period of growth.
We felt that there was little
information conveyed by the Inception Score that was not
conveyed by the Frechet Inception Distance, so for reasons
of space we have put the Inception Score plots in the supplementary
material.
}
\label{fig:cifar}
\end{figure*}

\begin{figure*}[ht]
\includegraphics[width=1.0\textwidth]{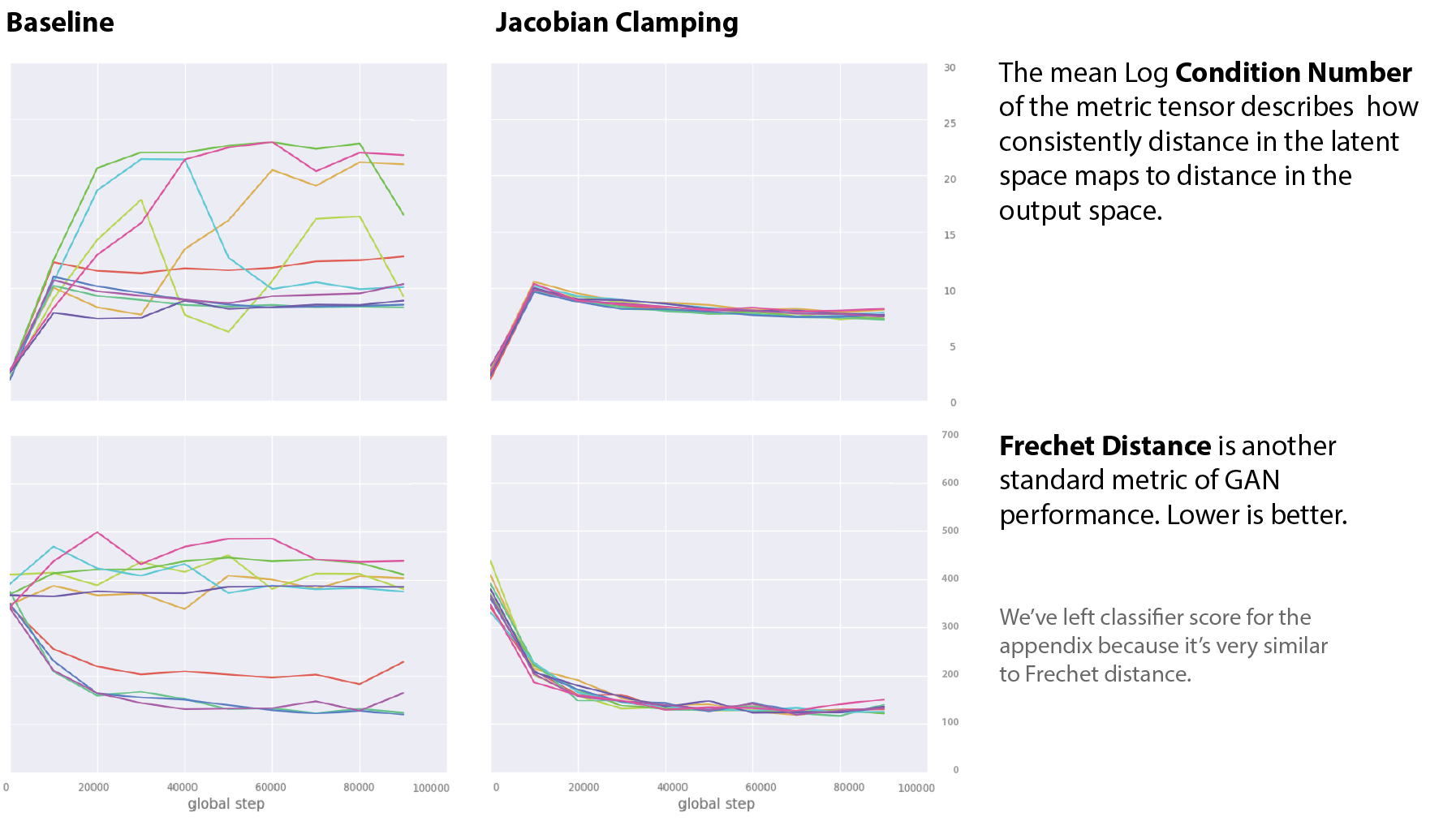}
\caption{
STL10 Experiments. Left and right columns correspond to 10 runs without and with Jacobian Clamping, respectively.
Within each column, each run has a unique color.
Top Row: Mean log-condition number over time. Bottom Row: Frechet Inception Distance over time. Note the red run (Left): the generator has
a condition number between the ill-conditioned cluster and the well-conditioned one; it also has scores between the low-scoring cluster
and the high-scoring one. Note also the light purple run (Left) which is similar.
As in Figure \ref{fig:cifar}, we have moved the Inception Score plots to the supplementary material.
}
\label{fig:stl}
\end{figure*}

\paragraph{Description of the Jacobian Clamping Technique:}
The technique we propose here is the simplest technique that we could get
working. We tried other more complicated techniques, but they did not perform
substantially better. An informal description is as follows:
We feed 2 mini-batches at a time to the generator. One batch is noise sampled
from $p_z$, the other is identical to the first but with small perturbations
added. The size of the perturbations is governed by a hyperparameter $\epsilon$.
We then take the norm of the change in outputs from batch to batch
and divide it by the norm of the change in inputs from batch to batch and apply
a penalty if that quotient becomes larger than some chosen hyperparameter
$\lambda_{max}$ or smaller than another hyperparameter $\lambda_{min}$.
The rough effect of this technique should be to encourage all of the
singular values of $J_z$ to lie within $[\lambda_{min}, \lambda_{max}]$ for all
$z$. See Algorithm \ref{alg:clamping} for a more formal description.

With respect to the goal of performing an intervention study, Jacobian Clamping
is slightly flawed because it does not directly penalize the condition number.
Unfortunately, directly penalizing the condition number during training is not
straightforward due to issues efficiently estimating the smallest eigenvalue
\citep{NUMERICALLINEARALGEBRA}. We choose not to worry about this too much;
We are more interested in understanding how the
spectrum of $J_z$ influences GAN training than in whether the
condition number is precisely the right summary quantity to be thinking about.

\begin{algorithm}[tb]
   \caption{Jacobian Clamping}
   \label{alg:clamping}
\begin{algorithmic}
   \STATE {\bfseries Input:} norm $\epsilon$, target quotients $\lambda_{max}, \lambda_{min}$, batch size $B$
   \REPEAT
   \STATE $z \in R^{B \times n_z} \sim p_z$.
   \STATE $\delta \in R^{B \times n_z} \sim N(0,1)$.
   \STATE $\delta := (\delta / ||\delta||) \epsilon$
   \STATE $z' := z + \delta$
   \STATE $Q := ||G(z) - G(z')|| / ||z - z'||$
   \STATE $L_{max} = (\max(Q, \lambda_{max}) - \lambda_{max})^2$
   \STATE $L_{min} = (\min(Q, \lambda_{min}) - \lambda_{min})^2$
   \STATE $L = L_{max} + L_{min} $
   \STATE Perform normal GAN update on $z$ with $L$ added to generator loss.
   \UNTIL{Training Finished}
\end{algorithmic}
\end{algorithm}

\paragraph{Jacobian Clamping Improves Mean Score and Reduces Variance of
Scores:}
In this section we evaluate the effects of using Jacobian Clamping.
Our aim here is not to make claims of State-of-the-Art scores\footnote{
We regard these claims as problematic anyway. One issue (among many) is that
scores are often reported from a single run, while the improvement in score
associated with a given method tends to be of the same scale as the inter-run
variance in scores.} but to provide evidence of a causal relationship between
the spectrum of $M_z$ and the scores.
Jacobian Clamping directly controls the conditition number of $M_z$.
We show (across 3 standard datasets) that when we implement Jacobian Clamping,
the condition number of the generator is decreased, and there is a corresponding
improvement in the quality of the scores.
This is evidence in favor of the hypothesis that ill-conditioning of $M_z$
``causes'' bad scores.

Specifically, we train the same models as from the previous section using
Jacobian Clamping with a $\lambda_{max}$ of 20, a $\lambda_{min}$ of
1, and $\epsilon$ of 1 and hold everything else the same.
As in the previous section, we conducted 10 training runs for each dataset.
Broadly speaking, the effect of Jacobian Clamping was to prevent the GANs from
falling into the ill-conditioned cluster.
This improved the average case performance, but didn't improve the
best case performance.
For all 3 datasets, we show terminal log spectra of $E_z[J_z]$ in Figure
\ref{fig:spectra}.

We first discuss the MNIST results.
The right column of Figure \ref{fig:mnist} shows measurements from 10 runs
using Jacobian Clamping. As compared to their
``unregularized'' counterparts in the left column, the runs using Jacobian
Clamping all show condition numbers that stop growing early in training.
The runs using Jacobian Clamping have scores similar to the best scores
achieved by runs without.
The scores also show lower intra-run variance for the ``regularized runs''.

The story is similar for CIFAR-10 and STL-10, the results for which can be
seen in the right columns of Figures \ref{fig:cifar} and \ref{fig:stl}
respectively.
For CIFAR-10, 9 out of 10 runs using Jacobian Clamping fell into the
``good cluster''.
The run that scored poorly also had a generator with a high condition number.
\textbf{It is noteworthy that the failure mode  we observed was one in
which the technique failed to constrain the quotient $Q$ rather than one in
which the quotient $Q$ was constrained and failure occured anyway.}
It is also (weak) evidence in favor of the causality hypothesis (in particular,
it is evidence against the alternative hypothesis that Jacobian Clamping acts
to increase scores in some other way than by constraining the conditioning).
For STL-10, all runs fell into the good cluster.

It's worth mentioning how we chose the values of the hyperparameters:
For $\epsilon$ we chose a value of 1 and never changed it because it seemed
to work well enough.
We then looked at the empirical value of the quotient $Q$ from Algorithm
\ref{alg:clamping} during training without Jacobian Clamping.
We set $\lambda_{min}, \lambda_{max}$ such that the
runs that achieved good scores had $Q$ mostly lying between those two values.
We consider the ability to perform this procedure an advantage of Jacobian
Clamping. Most techniques that introduce hyperparameters don't come bundled
with an algorithm to automatically set those hyperparameters.

We have observed that intervening to improve generator conditioning improves
generator performance during GAN training. In the supplementary material,
we discuss whether this
relationship between conditioning and performance holds for all possible
generators.

\begin{figure}[ht]
\vskip 0.2in
\begin{center}
\centerline{\includegraphics[width=\columnwidth]{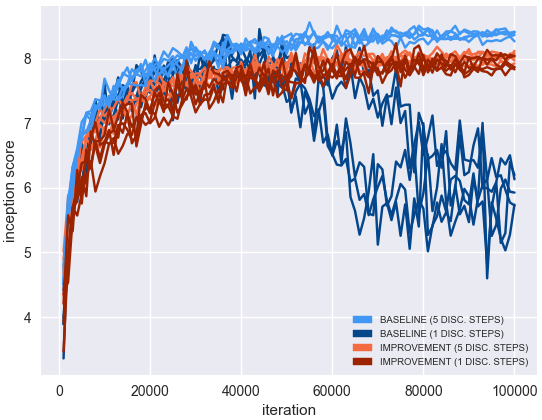}}
\caption{
We train a class-conditional WGAN-GP under 4 different settings:
With and without Jacobian Clamping and with 5 and 1 discriminator updates.
We perform 5 trials for each setting.
We made no effort whatsoever to tune our hyperparameters.
We find that when using only 1 discriminator update the baseline model
``collapses'' but the model with Jacobian Clamping does not.
When using 5 discriminator updates, the model with Jacobian Clamping performs
slightly worse in terms of Inception Score, but this difference is small and
could easily be due to the fact that we did not perform any tuning.
When using Jacobian Clamping, reducing the number of discriminator steps
does not reduce the score, but it more than halves the wall-clock time.
}
\label{fig:wgangp}
\end{center}
\vskip -0.2in
\end{figure}

\paragraph{Jacobian Clamping Speeds Up State-of-the-Art Models:}
One limitation of the experimental results we've discussed so far is that they
were obtained on a baseline model that does not include modifications that have
very recently become popular in the GAN literature.
We would like to know how Jacobian Clamping interacts with such modifications
as the Wasserstein loss \citep{WGAN}, the gradient penalty \citep{WGANGP}, and
various methods of conditioning the generator on label information
\citep{CONDITIONAL,ACGAN}.
Exhaustively evaluating all of these combinations is outside the scope of this
work, so we chose one existing implementation to assess the generality of
our findings.

We use the software implementation of a conditional GAN with gradient
penalty from
\url{https://github.com/igul222/improved_wgan_training} as our baseline because
this is the model from \citet{WGANGP} that scored the highest.
With its default hyperparameters this model has little variance in scores
between runs but is quite slow, as it performs 5 discriminator updates per
generator update.
It would thus be desirable to find a way to achieve the same results with
fewer discriminator updates.
Loosely following \citet{CRAMERGAN}, we jointly vary the number of discriminator steps
and whether Jacobian Clamping is applied.
\textbf{Using the same hyperparameters as in previous experiments (that is, we made no
attempt to tune for score) we find that reducing the number of discriminator
updates and adding Jacobian Clamping more than halves the wall-clock time
with little degradation in score.}
See Figure \ref{fig:wgangp} for more details.

\section{Related Work}
\label{section:relatedwork}

\textbf{GANs and other Deep Generative Models:}
There has been too much work on
GANs to adequately survey it here,
so we give an incomplete sketch:
One strand attempts to scale GANs up to work on larger datasets of
high resolution images with more variability
\cite{LAPGAN,DCGAN,ACGAN,STACKGAN,PROGRESSIVEGAN,PROJECTION,SAGAN}.
Yet another focuses on applications such as image-to-image translation
\cite{CYCLEGAN}, domain adaptation \cite{DOMAINADAPTATION}, and
super-resolution \cite{SUPERRESOLUTION}.
Other work focuses on addressing pathologies of the training
procedure \cite{UNROLLEDGAN}, on making theoretical claims
\cite{GENERALIZATIONANDEQUILIBRIUM}
or on evaluating trained GANS
\cite{LEARNTHEDISTRIBUTION}.
In spectral normalization \cite{SPECTRALNORM},
the largest singular value of the individual
layer Jacobians in the discriminator is approximately penalized using the power
method (see \citet{NUMERICALLINEARALGEBRA} for an explanation of this).
If Jacobian Clamping is performed with $\lambda_{min} = 0$, then it is vaguely
similar to performing spectral normalization \textit{on the generator}.
See \citet{GANSURVEY} for a more full accounting.

\textbf{Geometry and Neural Networks:}
Early work on geometry and neural networks
includes the Contractive Autoencoder
\cite{CONTRACTIVEAUTOENCODERS} in which an autoencoder is modified by penalizing
norm of the derivatives of its hidden units with respect to its input.
\citet{REPRESENTATIONLEARNING} discuss an
interpretation of representation learning as manifold learning.
More recently, \citet{MANIFOLDGAN} improved semi-supervised learning results
by enforcing geometric invariances on the classifier and
\citet{DYNAMICALISOMETRY} study the spectrum of squared singular values of the
input-output Jacobian for feed-forward classifiers with random weights.
\citet{SENSITIVITYANDGENERALIZATION} explore the relationship between the norm
of that Jacobian and the generalization error of the classifier.
In a related vein, three similar papers \cite{ODDITY,METRICS,GEOMETRY} have
explicitly studied variational autoencoders through the lens of geometry.

\textbf{Invertible Density Estimators and Adversarial Training:}
In \citet{FLOWGAN} and \citet{NVPGAN}, adversarial training is compared to
maximum likelihood training of generative image models using
an invertible decoder as in \citet{NICE,nvp}.
They find that the decoder spectrum drops off more quickly when using
adversarial training than when using maximum likelihood training.
This finding is evidence that ill-conditioning of the generator is somehow
fundamentally coupled with adversarial training techniques.
Our work instead studies the variation of the conditioning among many runs of
the same GAN training procedure, going on to show that this variation
corresponds to the variation in scores and that intervening with Jacobian
Clamping dramatically changes this variation.
We also find that the ill-conditioning does not always happen for adversarial
training --- see Figure \ref{fig:spectra}.

\section{Conclusions and Future Work}
\label{section:conclusion}

We studied the dynamics of the generator Jacobian and found that (during
training) it generally becomes ill-conditioned everywhere.
We then noted a strong correspondence between the conditioning of the Jacobian
and two quantitative metrics for evaluating GANs.
By explicitly controlling the conditioning during training
through a technique that we call Jacobian Clamping, we were able to improve
the two other quantitative measures of GAN performance.
We thus provided evidence that there is a causal relationship between the
conditioning of GAN generators and the ``quality'' of the models represented
by those GAN generators.
We believe this work represents a significant step toward understanding
GAN training dynamics.

\section*{Acknowledgements}
We thank Ben Poole, Luke Metz, Jonathon Shlens, Vincent Dumoulin and
Balaji Lakshminarayanan for commenting on earlier drafts.
We thank Ishaan Gulrajani for sharing code for a baseline CIFAR-10
GAN implementation.
We thank Daniel Duckworth for help implementing an efficient
Jacobian computation in TensorFlow.
We thank Sam Schoenholz, Matt Hoffman, Nic Ford, Jascha Sohl-Dickstein,
Justin Gilmer, George Dahl, and Matthew Johnson for helpful discussions
regarding the content of this work.

\bibliography{paper}

\begin{thebibliography}{48}
\providecommand{\natexlab}[1]{#1}
\providecommand{\url}[1]{\texttt{#1}}
\expandafter\ifx\csname urlstyle\endcsname\relax
  \providecommand{\doi}[1]{doi: #1}\else
  \providecommand{\doi}{doi: \begingroup \urlstyle{rm}\Url}\fi

\bibitem[{Arjovsky} \& {Bottou}(2017){Arjovsky} and {Bottou}]{PRINCIPLED}
{Arjovsky}, M. and {Bottou}, L.
\newblock {Towards Principled Methods for Training Generative Adversarial
  Networks}.
\newblock \emph{ArXiv e-prints}, January 2017.

\bibitem[{Arjovsky} et~al.(2017){Arjovsky}, {Chintala}, and {Bottou}]{WGAN}
{Arjovsky}, M., {Chintala}, S., and {Bottou}, L.
\newblock {Wasserstein GAN}.
\newblock \emph{ArXiv e-prints}, January 2017.

\bibitem[Arora \& Zhang(2017)Arora and Zhang]{LEARNTHEDISTRIBUTION}
Arora, S. and Zhang, Y.
\newblock Do gans actually learn the distribution? an empirical study.
\newblock \emph{CoRR}, abs/1706.08224, 2017.
\newblock URL \url{http://arxiv.org/abs/1706.08224}.

\bibitem[Arora et~al.(2017)Arora, Ge, Liang, Ma, and
  Zhang]{GENERALIZATIONANDEQUILIBRIUM}
Arora, S., Ge, R., Liang, Y., Ma, T., and Zhang, Y.
\newblock Generalization and equilibrium in generative adversarial nets (gans).
\newblock \emph{CoRR}, abs/1703.00573, 2017.
\newblock URL \url{http://arxiv.org/abs/1703.00573}.

\bibitem[{Arvanitidis} et~al.(2017){Arvanitidis}, {Hansen}, and
  {Hauberg}]{ODDITY}
{Arvanitidis}, G., {Hansen}, L.~K., and {Hauberg}, S.
\newblock {Latent Space Oddity: on the Curvature of Deep Generative Models}.
\newblock \emph{ArXiv e-prints}, October 2017.

\bibitem[Bellemare et~al.(2017)Bellemare, Danihelka, Dabney, Mohamed,
  Lakshminarayanan, Hoyer, and Munos]{CRAMERGAN}
Bellemare, M.~G., Danihelka, I., Dabney, W., Mohamed, S., Lakshminarayanan, B.,
  Hoyer, S., and Munos, R.
\newblock The cramer distance as a solution to biased wasserstein gradients.
\newblock \emph{CoRR}, abs/1705.10743, 2017.
\newblock URL \url{http://arxiv.org/abs/1705.10743}.

\bibitem[Bengio et~al.(2012)Bengio, Courville, and
  Vincent]{REPRESENTATIONLEARNING}
Bengio, Y., Courville, A.~C., and Vincent, P.
\newblock Unsupervised feature learning and deep learning: {A} review and new
  perspectives.
\newblock \emph{CoRR}, abs/1206.5538, 2012.
\newblock URL \url{http://arxiv.org/abs/1206.5538}.

\bibitem[Bousmalis et~al.(2016)Bousmalis, Silberman, Dohan, Erhan, and
  Krishnan]{DOMAINADAPTATION}
Bousmalis, K., Silberman, N., Dohan, D., Erhan, D., and Krishnan, D.
\newblock Unsupervised pixel-level domain adaptation with generative
  adversarial networks.
\newblock \emph{CoRR}, abs/1612.05424, 2016.
\newblock URL \url{http://arxiv.org/abs/1612.05424}.

\bibitem[{Chen} et~al.(2017){Chen}, {Klushyn}, {Kurle}, {Jiang}, {Bayer}, and
  {van der Smagt}]{METRICS}
{Chen}, N., {Klushyn}, A., {Kurle}, R., {Jiang}, X., {Bayer}, J., and {van der
  Smagt}, P.
\newblock {Metrics for Deep Generative Models}.
\newblock \emph{ArXiv e-prints}, November 2017.

\bibitem[Coates et~al.(2011)Coates, Ng, and Lee]{STL}
Coates, A., Ng, A., and Lee, H.
\newblock An analysis of single-layer networks in unsupervised feature
  learning.
\newblock In \emph{Proceedings of the fourteenth international conference on
  artificial intelligence and statistics}, pp.\  215--223, 2011.

\bibitem[Danihelka et~al.(2017)Danihelka, Lakshminarayanan, Uria, Wierstra, and
  Dayan]{NVPGAN}
Danihelka, I., Lakshminarayanan, B., Uria, B., Wierstra, D., and Dayan, P.
\newblock Comparison of maximum likelihood and gan-based training of real nvps.
\newblock \emph{CoRR}, abs/1705.05263, 2017.
\newblock URL \url{http://arxiv.org/abs/1705.05263}.

\bibitem[Denton et~al.(2015)Denton, Chintala, Szlam, and Fergus]{LAPGAN}
Denton, E.~L., Chintala, S., Szlam, A., and Fergus, R.
\newblock Deep generative image models using a laplacian pyramid of adversarial
  networks.
\newblock \emph{CoRR}, abs/1506.05751, 2015.
\newblock URL \url{http://arxiv.org/abs/1506.05751}.

\bibitem[Dinh et~al.(2014)Dinh, Krueger, and Bengio]{NICE}
Dinh, L., Krueger, D., and Bengio, Y.
\newblock {NICE:} non-linear independent components estimation.
\newblock \emph{CoRR}, abs/1410.8516, 2014.
\newblock URL \url{http://arxiv.org/abs/1410.8516}.

\bibitem[Dinh et~al.(2016)Dinh, Sohl{-}Dickstein, and Bengio]{nvp}
Dinh, L., Sohl{-}Dickstein, J., and Bengio, S.
\newblock Density estimation using real {NVP}.
\newblock \emph{CoRR}, abs/1605.08803, 2016.
\newblock URL \url{http://arxiv.org/abs/1605.08803}.

\bibitem[Eberhardt \& Scheines(2007)Eberhardt and
  Scheines]{INTERVENTIONSANDCAUSALINFERENCE}
Eberhardt, F. and Scheines, R.
\newblock Interventions and causal inference.
\newblock \emph{Philosophy of Science}, 74\penalty0 (5):\penalty0 981--995,
  2007.

\bibitem[Golub \& Van~Loan(1996)Golub and Van~Loan]{NUMERICALLINEARALGEBRA}
Golub, G.~H. and Van~Loan, C.~F.
\newblock \emph{Matrix Computations (3rd Ed.)}.
\newblock Johns Hopkins University Press, Baltimore, MD, USA, 1996.
\newblock ISBN 0-8018-5414-8.

\bibitem[{Goodfellow}(2017)]{GANSURVEY}
{Goodfellow}, I.
\newblock {NIPS 2016 Tutorial: Generative Adversarial Networks}.
\newblock \emph{ArXiv e-prints}, December 2017.

\bibitem[{Goodfellow} et~al.(2014){Goodfellow}, {Pouget-Abadie}, {Mirza}, {Xu},
  {Warde-Farley}, {Ozair}, {Courville}, and {Bengio}]{GANS}
{Goodfellow}, I.~J., {Pouget-Abadie}, J., {Mirza}, M., {Xu}, B.,
  {Warde-Farley}, D., {Ozair}, S., {Courville}, A., and {Bengio}, Y.
\newblock {Generative Adversarial Networks}.
\newblock \emph{ArXiv e-prints}, June 2014.

\bibitem[Grover et~al.(2017)Grover, Dhar, and Ermon]{FLOWGAN}
Grover, A., Dhar, M., and Ermon, S.
\newblock Flow-gan: Bridging implicit and prescribed learning in generative
  models.
\newblock \emph{CoRR}, abs/1705.08868, 2017.
\newblock URL \url{http://arxiv.org/abs/1705.08868}.

\bibitem[Gulrajani et~al.(2017)Gulrajani, Ahmed, Arjovsky, Dumoulin, and
  Courville]{WGANGP}
Gulrajani, I., Ahmed, F., Arjovsky, M., Dumoulin, V., and Courville, A.~C.
\newblock Improved training of wasserstein gans.
\newblock \emph{CoRR}, abs/1704.00028, 2017.
\newblock URL \url{http://arxiv.org/abs/1704.00028}.

\bibitem[Hagmayer et~al.(2007)Hagmayer, Sloman, Lagnado, and
  Waldmann]{CAUSALREASONINGTHROUGHINTERVENTION}
Hagmayer, Y., Sloman, S.~A., Lagnado, D.~A., and Waldmann, M.~R.
\newblock Causal reasoning through intervention.
\newblock \emph{Causal learning: Psychology, philosophy, and computation}, pp.\
   86--100, 2007.

\bibitem[{Heusel} et~al.(2017){Heusel}, {Ramsauer}, {Unterthiner}, {Nessler},
  and {Hochreiter}]{FID}
{Heusel}, M., {Ramsauer}, H., {Unterthiner}, T., {Nessler}, B., and
  {Hochreiter}, S.
\newblock {GANs Trained by a Two Time-Scale Update Rule Converge to a Local
  Nash Equilibrium}.
\newblock \emph{ArXiv e-prints}, June 2017.

\bibitem[Hoffman(2017)]{MCMCVAE}
Hoffman, M.~D.
\newblock Learning deep latent gaussian models with markov chain monte carlo.
\newblock In \emph{International Conference on Machine Learning}, pp.\
  1510--1519, 2017.

\bibitem[{Karras} et~al.(2017){Karras}, {Aila}, {Laine}, and
  {Lehtinen}]{PROGRESSIVEGAN}
{Karras}, T., {Aila}, T., {Laine}, S., and {Lehtinen}, J.
\newblock {Progressive Growing of GANs for Improved Quality, Stability, and
  Variation}.
\newblock \emph{ArXiv e-prints}, October 2017.

\bibitem[{Kingma} \& {Welling}(2013){Kingma} and {Welling}]{VAES}
{Kingma}, D.~P. and {Welling}, M.
\newblock {Auto-Encoding Variational Bayes}.
\newblock \emph{ArXiv e-prints}, December 2013.

\bibitem[Krizhevsky(2009)]{CIFAR}
Krizhevsky, A.
\newblock Learning multiple layers of features from tiny images.
\newblock 2009.

\bibitem[{Kumar} et~al.(2017){Kumar}, {Sattigeri}, and {Fletcher}]{MANIFOLDGAN}
{Kumar}, A., {Sattigeri}, P., and {Fletcher}, P.~T.
\newblock {Improved Semi-supervised Learning with GANs using Manifold
  Invariances}.
\newblock \emph{ArXiv e-prints}, May 2017.

\bibitem[LeCun et~al.(1998)LeCun, Bottou, Bengio, and Haffner]{MNIST}
LeCun, Y., Bottou, L., Bengio, Y., and Haffner, P.
\newblock Gradient-based learning applied to document recognition.
\newblock \emph{Proceedings of the IEEE}, 1998.

\bibitem[{Ledig} et~al.(2016){Ledig}, {Theis}, {Huszar}, {Caballero}, {Aitken},
  {Tejani}, {Totz}, {Wang}, and {Shi}]{SUPERRESOLUTION}
{Ledig}, C., {Theis}, L., {Huszar}, F., {Caballero}, J., {Aitken}, A.,
  {Tejani}, A., {Totz}, J., {Wang}, Z., and {Shi}, W.
\newblock {Photo-Realistic Single Image Super-Resolution Using a Generative
  Adversarial Network}.
\newblock \emph{ArXiv e-prints}, September 2016.

\bibitem[Metz et~al.(2016)Metz, Poole, Pfau, and Sohl{-}Dickstein]{UNROLLEDGAN}
Metz, L., Poole, B., Pfau, D., and Sohl{-}Dickstein, J.
\newblock Unrolled generative adversarial networks.
\newblock \emph{CoRR}, abs/1611.02163, 2016.
\newblock URL \url{http://arxiv.org/abs/1611.02163}.

\bibitem[Mirza \& Osindero(2014)Mirza and Osindero]{CONDITIONAL}
Mirza, M. and Osindero, S.
\newblock Conditional generative adversarial nets.
\newblock \emph{CoRR}, abs/1411.1784, 2014.
\newblock URL \url{http://arxiv.org/abs/1411.1784}.

\bibitem[Miyato \& Koyama(2018)Miyato and Koyama]{PROJECTION}
Miyato, T. and Koyama, M.
\newblock c{GAN}s with projection discriminator.
\newblock In \emph{International Conference on Learning Representations}, 2018.
\newblock URL \url{https://openreview.net/forum?id=ByS1VpgRZ}.

\bibitem[Miyato et~al.(2018)Miyato, Kataoka, Koyama, and Yoshida]{SPECTRALNORM}
Miyato, T., Kataoka, T., Koyama, M., and Yoshida, Y.
\newblock Spectral normalization for generative adversarial networks.
\newblock In \emph{International Conference on Learning Representations}, 2018.
\newblock URL \url{https://openreview.net/forum?id=B1QRgziT-}.

\bibitem[Nagarajan \& Kolter(2017)Nagarajan and Kolter]{LOCALLYSTABLE}
Nagarajan, V. and Kolter, J.~Z.
\newblock Gradient descent {GAN} optimization is locally stable.
\newblock \emph{CoRR}, abs/1706.04156, 2017.
\newblock URL \url{http://arxiv.org/abs/1706.04156}.

\bibitem[Novak et~al.(2018)Novak, Bahri, Abolafia, Pennington, and
  Sohl-Dickstein]{SENSITIVITYANDGENERALIZATION}
Novak, R., Bahri, Y., Abolafia, D.~A., Pennington, J., and Sohl-Dickstein, J.
\newblock Sensitivity and generalization in neural networks: an empirical
  study.
\newblock \emph{arXiv preprint arXiv:1802.08760}, 2018.

\bibitem[{Odena} et~al.(2016){Odena}, {Olah}, and {Shlens}]{ACGAN}
{Odena}, A., {Olah}, C., and {Shlens}, J.
\newblock {Conditional Image Synthesis With Auxiliary Classifier GANs}.
\newblock \emph{ArXiv e-prints}, October 2016.

\bibitem[Pearl(2009)]{CAUSALITY}
Pearl, J.
\newblock \emph{Causality}.
\newblock Cambridge university press, 2009.

\bibitem[Pennington et~al.(2017)Pennington, Schoenholz, and
  Ganguli]{DYNAMICALISOMETRY}
Pennington, J., Schoenholz, S., and Ganguli, S.
\newblock Resurrecting the sigmoid in deep learning through dynamical isometry:
  theory and practice.
\newblock In Guyon, I., Luxburg, U.~V., Bengio, S., Wallach, H., Fergus, R.,
  Vishwanathan, S., and Garnett, R. (eds.), \emph{Advances in Neural
  Information Processing Systems 30}, pp.\  4788--4798. Curran Associates,
  Inc., 2017.

\bibitem[Radford et~al.(2015)Radford, Metz, and Chintala]{DCGAN}
Radford, A., Metz, L., and Chintala, S.
\newblock Unsupervised representation learning with deep convolutional
  generative adversarial networks.
\newblock \emph{CoRR}, abs/1511.06434, 2015.
\newblock URL \url{http://arxiv.org/abs/1511.06434}.

\bibitem[{Rezende} et~al.(2014){Rezende}, {Mohamed}, and {Wierstra}]{VAES2}
{Rezende}, D., {Mohamed}, S., and {Wierstra}, D.
\newblock {Stochastic Backpropagation and Approximate Inference in Deep
  Generative Models}.
\newblock \emph{ArXiv e-prints}, January 2014.

\bibitem[Rifai et~al.(2011)Rifai, Vincent, Muller, Glorot, and
  Bengio]{CONTRACTIVEAUTOENCODERS}
Rifai, S., Vincent, P., Muller, X., Glorot, X., and Bengio, Y.
\newblock Contractive auto-encoders: Explicit invariance during feature
  extraction.
\newblock In \emph{Proceedings of the 28th international conference on machine
  learning (ICML-11)}, pp.\  833--840, 2011.

\bibitem[{Salimans} et~al.(2016){Salimans}, {Goodfellow}, {Zaremba}, {Cheung},
  {Radford}, and {Chen}]{IMPROVEDTECHNIQUES}
{Salimans}, T., {Goodfellow}, I., {Zaremba}, W., {Cheung}, V., {Radford}, A.,
  and {Chen}, X.
\newblock {Improved Techniques for Training GANs}.
\newblock \emph{ArXiv e-prints}, June 2016.

\bibitem[{Shao} et~al.(2017){Shao}, {Kumar}, and {Fletcher}]{GEOMETRY}
{Shao}, H., {Kumar}, A., and {Fletcher}, P.~T.
\newblock {The Riemannian Geometry of Deep Generative Models}.
\newblock \emph{ArXiv e-prints}, November 2017.

\bibitem[Woodward(2005)]{MAKINGTHINGSHAPPEN}
Woodward, J.
\newblock \emph{Making things happen: A theory of causal explanation}.
\newblock Oxford university press, 2005.

\bibitem[Wu et~al.(2016)Wu, Burda, Salakhutdinov, and Grosse]{AIS}
Wu, Y., Burda, Y., Salakhutdinov, R., and Grosse, R.~B.
\newblock On the quantitative analysis of decoder-based generative models.
\newblock \emph{CoRR}, abs/1611.04273, 2016.
\newblock URL \url{http://arxiv.org/abs/1611.04273}.

\bibitem[{Zhang} et~al.(2016){Zhang}, {Xu}, {Li}, {Zhang}, {Wang}, {Huang}, and
  {Metaxas}]{STACKGAN}
{Zhang}, H., {Xu}, T., {Li}, H., {Zhang}, S., {Wang}, X., {Huang}, X., and
  {Metaxas}, D.
\newblock {StackGAN: Text to Photo-realistic Image Synthesis with Stacked
  Generative Adversarial Networks}.
\newblock \emph{ArXiv e-prints}, December 2016.

\bibitem[Zhang et~al.(2018)Zhang, Goodfellow, Metaxas, and Odena]{SAGAN}
Zhang, H., Goodfellow, I., Metaxas, D., and Odena, A.
\newblock Self-attention generative adversarial networks.
\newblock \emph{arXiv preprint arXiv:1805.08318}, 2018.

\bibitem[{Zhu} et~al.(2017){Zhu}, {Park}, {Isola}, and {Efros}]{CYCLEGAN}
{Zhu}, J.-Y., {Park}, T., {Isola}, P., and {Efros}, A.~A.
\newblock {Unpaired Image-to-Image Translation using Cycle-Consistent
  Adversarial Networks}.
\newblock \emph{ArXiv e-prints}, March 2017.

\end{thebibliography}
\bibliographystyle{icml2018}

\clearpage

\appendix

\def \appverticalspace {-3ex}
\def \appsubfigurewidth {.45\textwidth}

\section{Why Compute the Condition Number?}
There are many summary statistics one could compute from the spectrum of the
Jacobian. It is not obvious a priori that it makes sense to focus on the ratio
of the maximum eigenvalue to the minimum eigenvalue, so here we make some
attempt to justify that decision.

If you were to just glance at the spectra figures provided in the main text,
using the log-determinant might seem like a reasonable thing to do.
However, we note that (at least for the MNIST experiments) the largest singular
values for the `well behaved' runs are distinctly lower than those for the
`poorly behaved' ones. This suggests that the conditioning might be more
pertinent than the determinant.

Even given that the conditioning is what's relevant, one could imagine other
measures of Jacobian conditioning that less strongly emphasize the extreme
singular values.
Indeed, computing such quantities would be a useful exercise, and we expect
that they would also correlate with GAN performance, but we have kept the
condition number because it is simple and well-understood.
We also feel that the condition number most closely relates to what is
being optimized by the Jacobian Clamping procedure.

\section{Additional Experimental Results}
\label{appendix}
This section contains results that we have included for the purpose of
completeness but which were not necessary for following the narrative of the
paper. References to this section can be found in the main text.

\subsection{Misbehaving Generators can be Well-Conditioned}
We have observed that intervening to improve generator
conditioning improves generator performance during
GAN training. We also might like to know whether this relationship holds for
all possible generators. Here we provide a counterexample
of a deliberately pathological generator (not trained with a GAN loss) which is
nonetheless well-conditioned. This suggests that the causal relationship we
explore in the main text may relate to the GAN training process,
and may not be an absolute property of generators in isolation.

\begin{figure}[ht]
\vskip 0.2in
\begin{center}
\centerline{\includegraphics[width=\columnwidth]{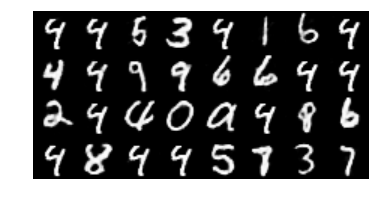}}
\caption{Samples from memorized $z$s. Half of the samples the generator was trained
to memorize are identical copies of a single MNIST sample (in this case, a four)
and the other half are random MNIST digits. The generator has successfully
memorized the $z$-to-digit associations it was trained to reproduce.}
\label{fig:half_samples_training_z}
\end{center}
\vskip -0.2in
\end{figure}

We train a generator using the DCGAN architecture with a latent
space of 64 dimensions. Rather than an adversarial loss, we train with an L2
reconstruction loss - in effect, teaching the generator to memorize the training
examples it has seen. We select 10,000 examples to memorize: half of them (5,000)
are random MNIST digits, and the other half are identical copies
of a single MNIST sample (in this case, a four). We then establish a consistent but
arbitrary mapping from 10,000 random $z$ values to the training examples.
The generator is trained with an L2 reconstruction loss to map each
memorized $z$ value to its associated training example. The generator's
behavior on non-memorized $z$ values is not considered at training time.
There is no discriminator involved in this training procedure. Figure~\ref{fig:half_samples_training_z}
shows the generator's output when provided the $z$ values it was trained
to associate with specific samples, indicating that it succeeds at memorizing
the half-random half-identical data it was trained on.

\begin{figure}[ht]
\vskip 0.2in
\begin{center}
\centerline{\includegraphics[width=\columnwidth]{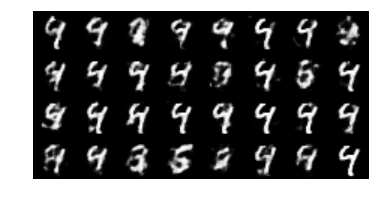}}
\caption{Samples from random $z$s. The generator's behavior on these $z$ values
was not considered at training time. These samples often resemble the
single four that made up half its training data, or other four- and
nine-like digits. Occasionally, it produces indistinct digits that are not
  four-like, such as the 3 and the 5 in the bottom row, or indistinct samples
that are not digit-like.}
\label{fig:half_samples_random_z}
\end{center}
\end{figure}

\begin{figure}[ht]
\vskip 0.2in
\begin{center}
\centerline{\includegraphics[width=\columnwidth]{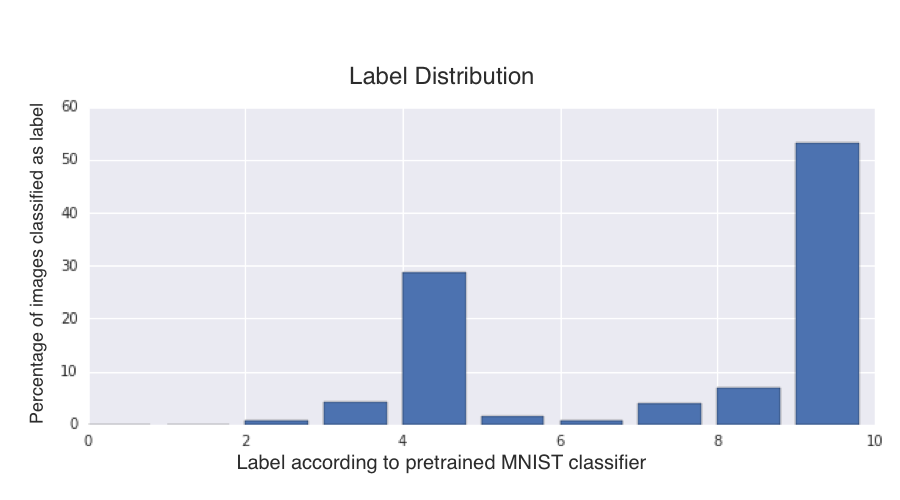}}
\caption{Label distribution of samples from random $z$s}
\label{fig:half_label_distribution_random_z}
\end{center}
\vskip -0.2in
\end{figure}

At evaluation time, we provide random latent vectors, rather than the latent
vectors the generator has been trained to memorize.
Figure~\ref{fig:half_samples_random_z}
shows the samples that this generator produces at evaluation time. This generator is
clearly not well-behaved: it suffers from mode collapse (i.e. it often
reproduces the single four that made up half of its training data) and mode dropping
(i.e. even when it produces a novel sample, it usually looks an indistinct four
or nine, and seldom looks like any other class).
Figure~\ref{fig:half_label_distribution_random_z} shows the label distribution
as measured by a pre-trained classifier, confirming that this generator
has a severe missing mode problem. This generator's poor behavior is also confirmed
by its scores. Its Classifier Score is 4.95 for memorized $z$ values
and 2.22 on non-memorized $z$ values. Its Frechet Distance is 118 for memorized
$z$ values and 240 for non-memorized $z$ values.

\begin{figure}[ht]
\vskip 0.2in
\begin{center}
\centerline{\includegraphics[width=\columnwidth]{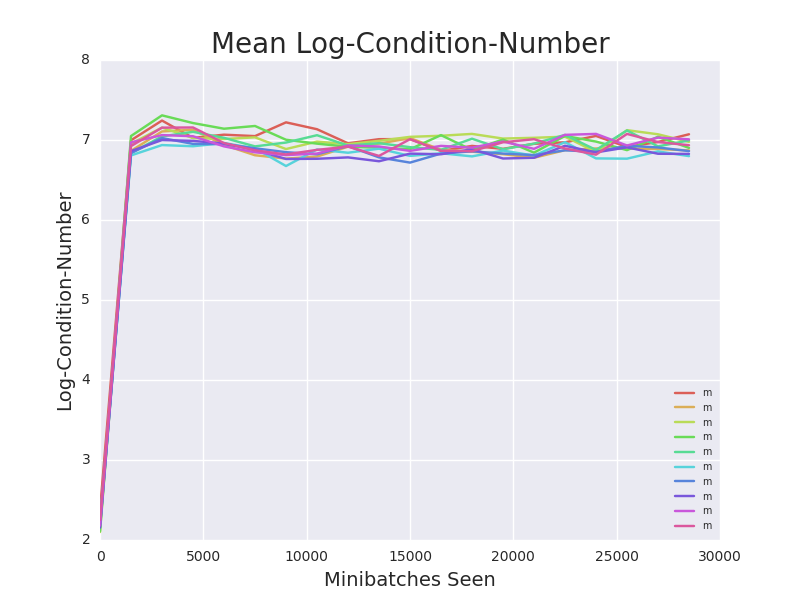}}
\caption{Mean log-condition number of misbehaving generator over 10 runs.
Compare to Figure~\ref{fig:mnist} in the main text: this misbehaving generator
is better-conditioned than the "good cluster" of GAN generators.}
\label{fig:half_log_condition_number}
\end{center}
\vskip -0.2in
\end{figure}

Figure~\ref{fig:half_log_condition_number} shows that this poorly-behaved generator
nonetheless has a good condition number. Taken in isolation, the trajectory of this
generator's condition number would suggest that it belongs in the "good cluster"
of Figure~\ref{fig:mnist}.

In summary, we demonstrate a generator that is \emph{not} trained with a GAN loss,
with conspicuous mode collapse and mode dropping, which is nonetheless
well-conditioned. This suggests that the relationship between generator
conditioning and generator performance does not hold for all generators,
and suggests that it may instead be a property of GAN training dynamics.

\vskip -0.2in

\begin{figure*}[ht]
\vspace{-0.2cm}
\begin{center}
  \begin{subfigure}[t]{\appsubfigurewidth}
    \centering
    \includegraphics[width=\textwidth]{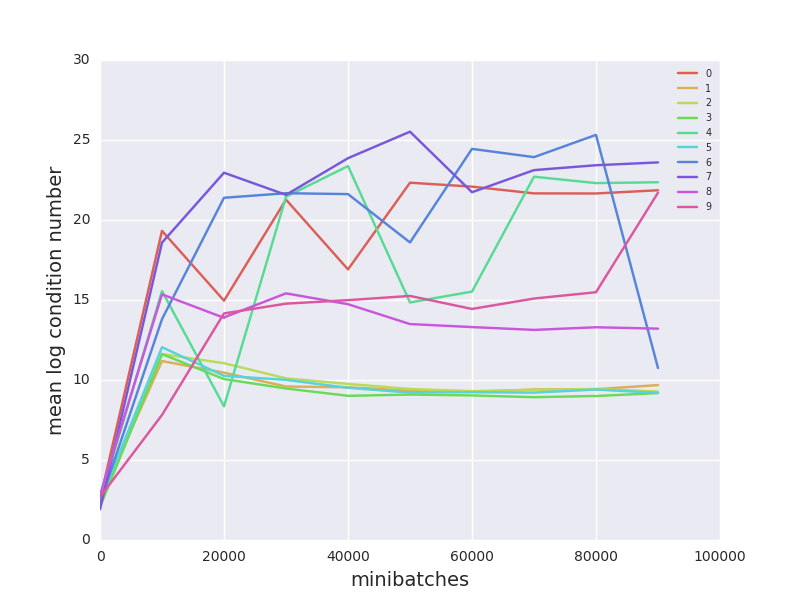}
    \label{cifar:a}
  \end{subfigure}
  \begin{subfigure}[t]{\appsubfigurewidth}
    \centering
    \includegraphics[width=\textwidth]{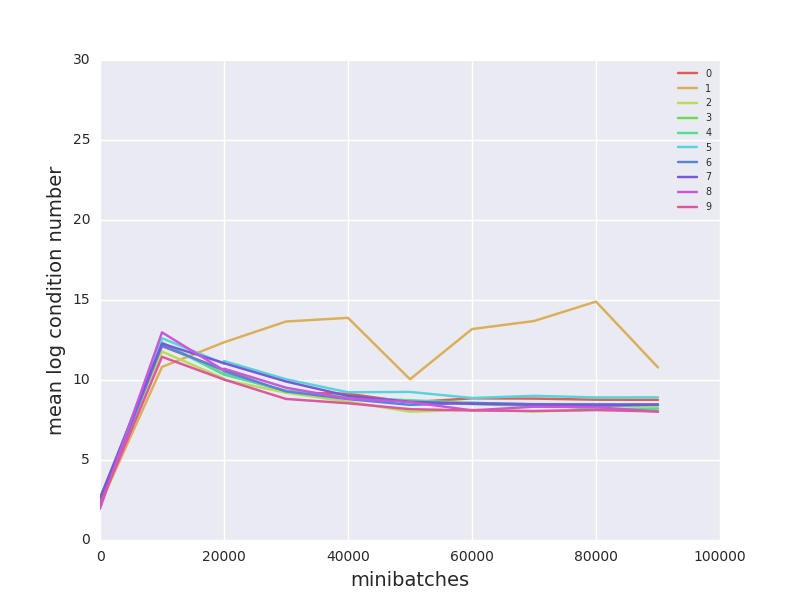}
    \label{cifar:b}
  \end{subfigure}\\[\appverticalspace]
  \begin{subfigure}[t]{\appsubfigurewidth}
    \centering
    \includegraphics[width=\textwidth]{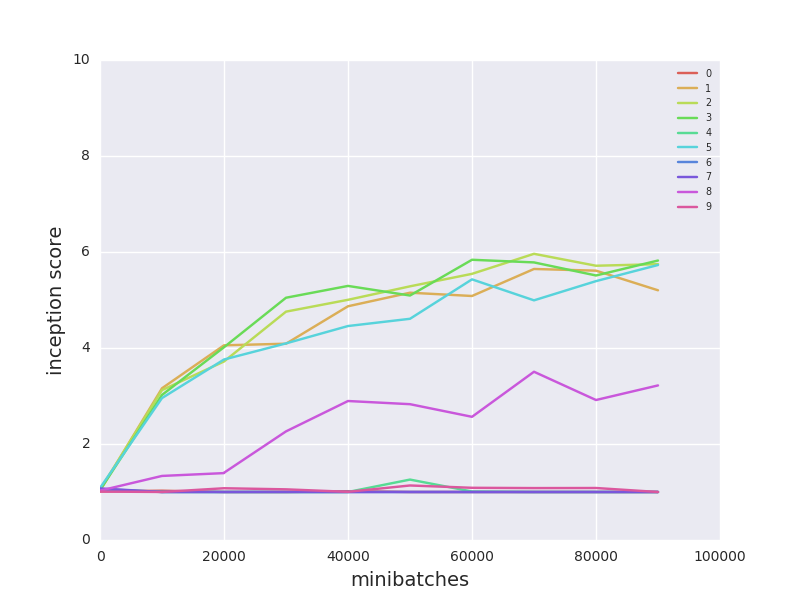}
    \label{cifar:c}
  \end{subfigure}
  \begin{subfigure}[t]{\appsubfigurewidth}
    \centering
    \includegraphics[width=\textwidth]{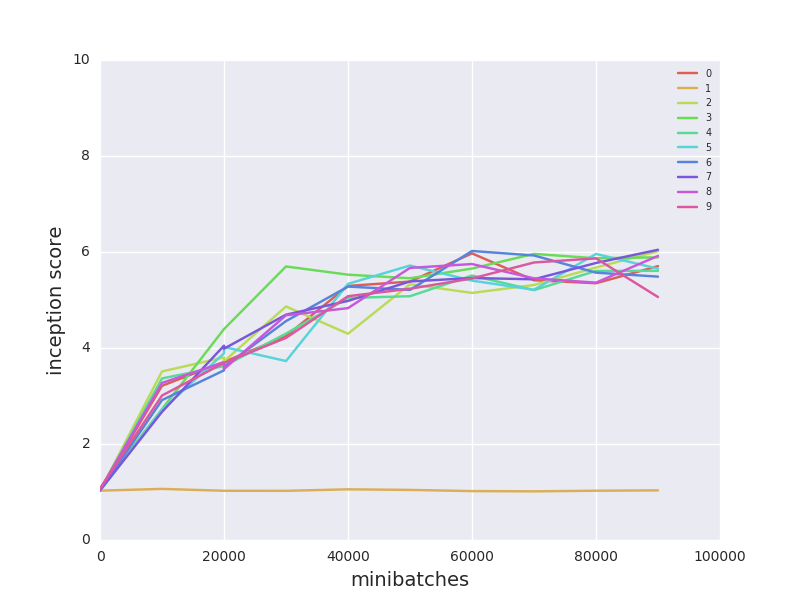}
    \label{cifar:d}
  \end{subfigure}\\[\appverticalspace]
  \begin{subfigure}[t]{\appsubfigurewidth}
    \centering
    \includegraphics[width=\textwidth]{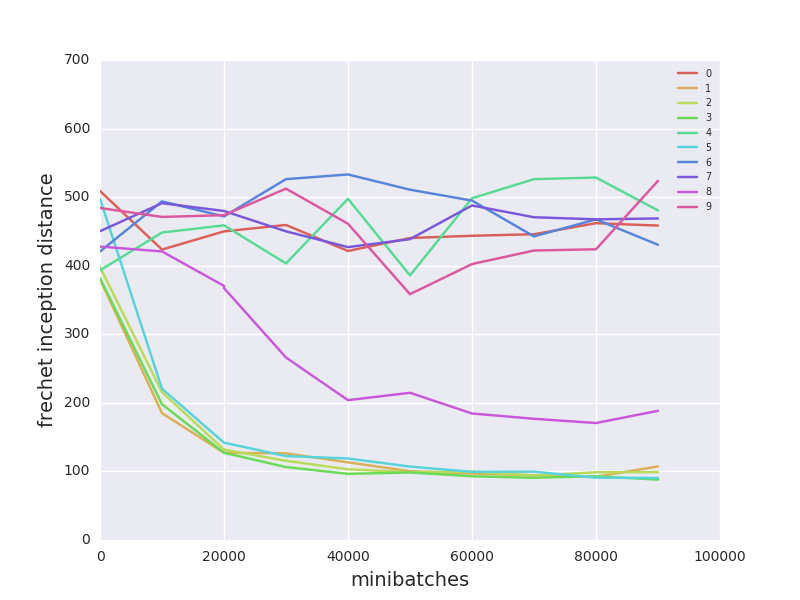}
    \label{cifar:e}
  \end{subfigure}
  \begin{subfigure}[t]{\appsubfigurewidth}
    \centering
    \includegraphics[width=\textwidth]{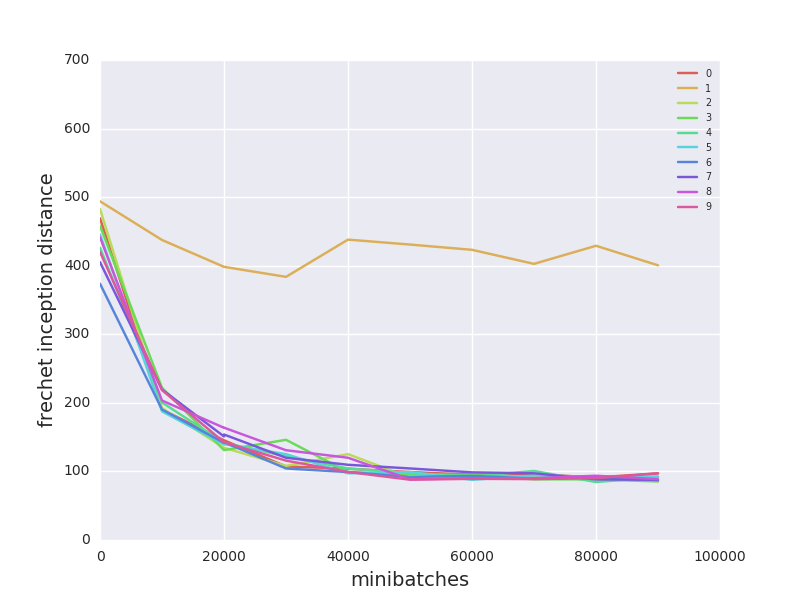}
    \label{cifar:f}
  \end{subfigure}\\[\appverticalspace]
  \begin{subfigure}[t]{\appsubfigurewidth}
    \centering
    \includegraphics[width=\textwidth]{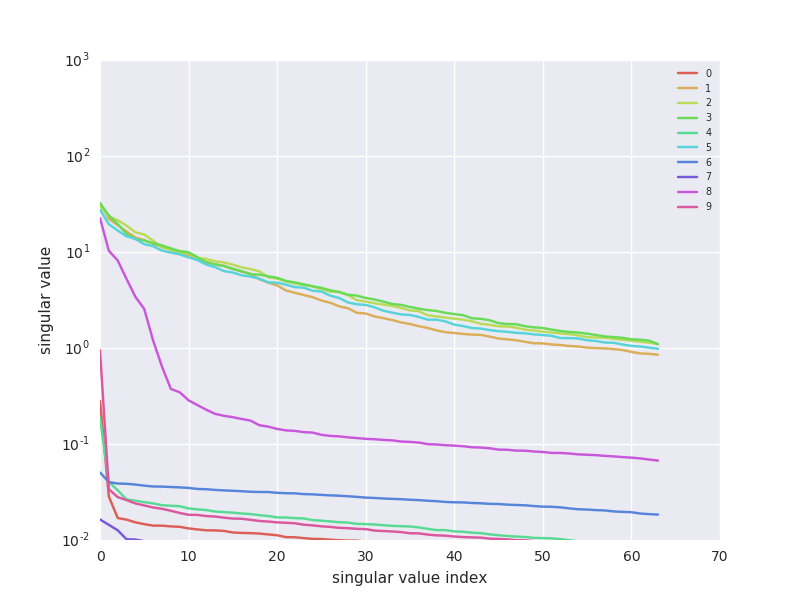}
    \label{cifar:e}
  \end{subfigure}
  \begin{subfigure}[t]{\appsubfigurewidth}
    \centering
    \includegraphics[width=\textwidth]{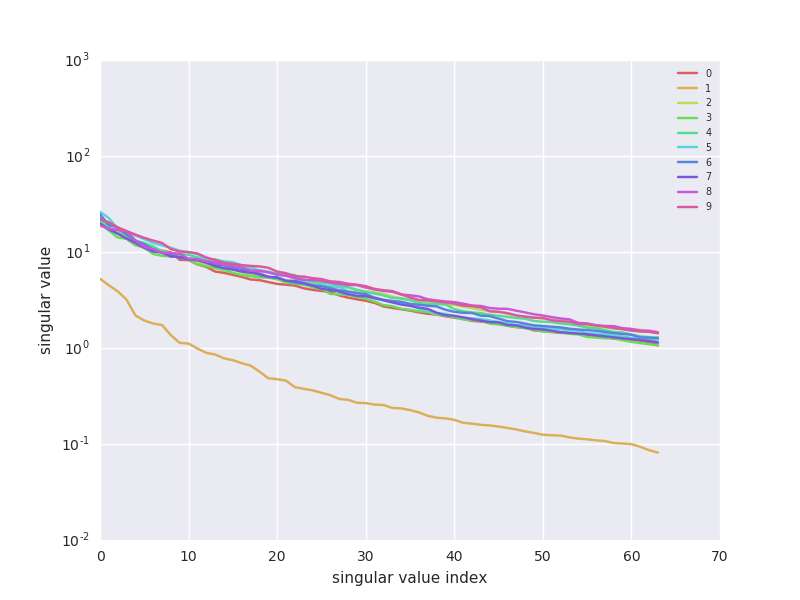}
    \label{cifar:f}
  \end{subfigure}
  \vspace{-0.8cm}
  \caption{
  CIFAR10 Experimental results. Left and right columns correspond to 10 runs without and with Jacobian Clamping, respectively. Within each column, each run has a unique color.
  }
  \label{fig:cifar}
\end{center}
\end{figure*}

\begin{figure*}[ht]
\vspace{-0.2cm}
\begin{center}
  \begin{subfigure}[t]{\appsubfigurewidth}
    \centering
    \includegraphics[width=\textwidth]{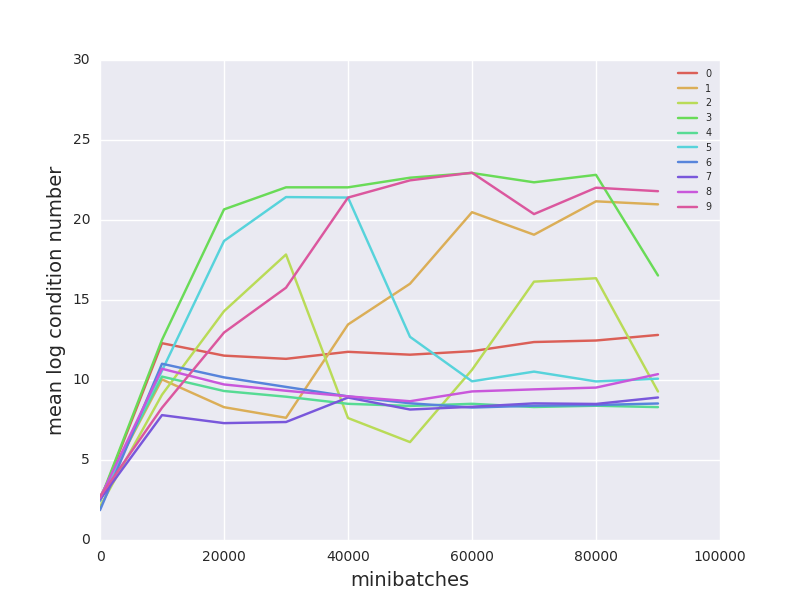}
    \label{stl:a}
  \end{subfigure}
  \begin{subfigure}[t]{\appsubfigurewidth}
    \centering
    \includegraphics[width=\textwidth]{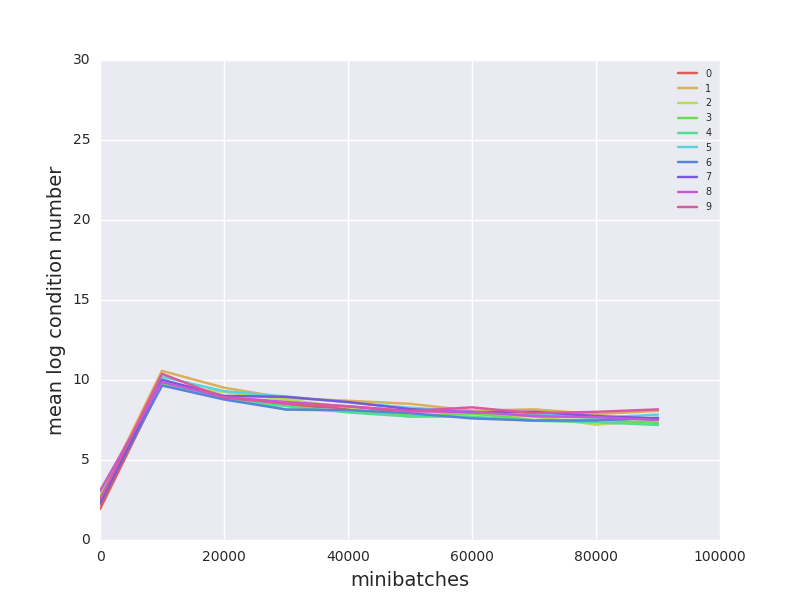}
    \label{stl:b}
  \end{subfigure}\\[\appverticalspace]
  \begin{subfigure}[t]{\appsubfigurewidth}
    \centering
    \includegraphics[width=\textwidth]{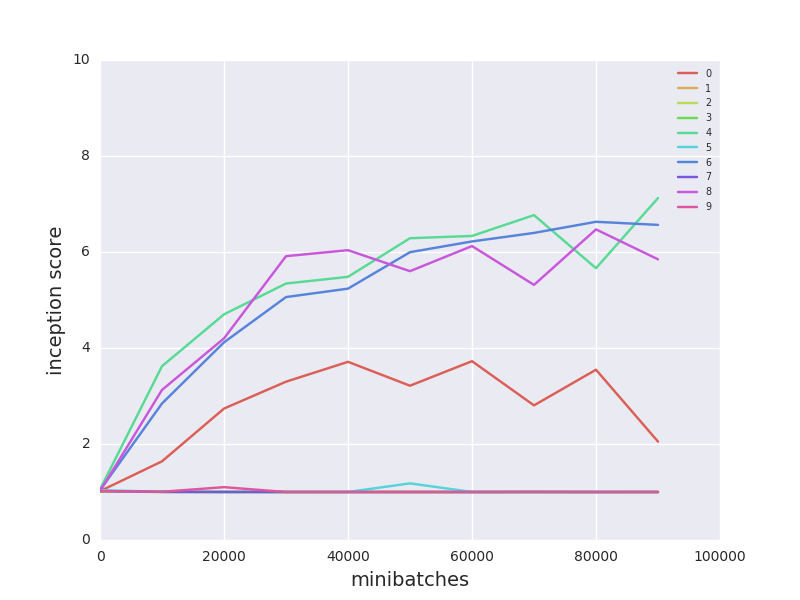}
    \label{stl:c}
  \end{subfigure}
  \begin{subfigure}[t]{\appsubfigurewidth}
    \centering
    \includegraphics[width=\textwidth]{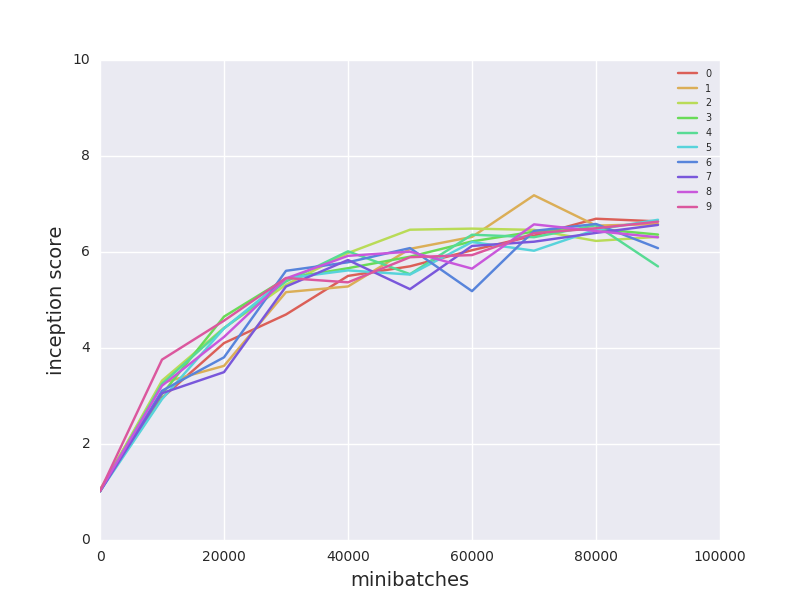}
    \label{stl:d}
  \end{subfigure}\\[\appverticalspace]
  \begin{subfigure}[t]{\appsubfigurewidth}
    \centering
    \includegraphics[width=\textwidth]{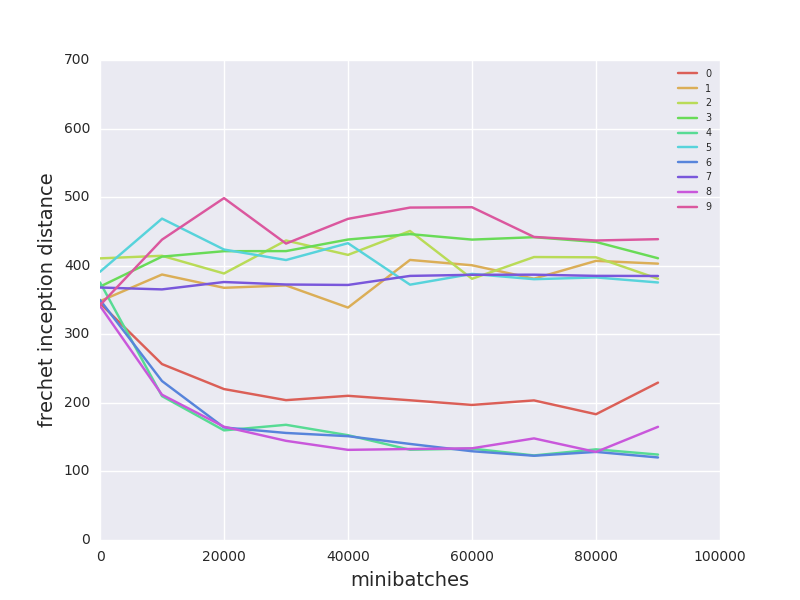}
    \label{stl:e}
  \end{subfigure}
  \begin{subfigure}[t]{\appsubfigurewidth}
    \centering
    \includegraphics[width=\textwidth]{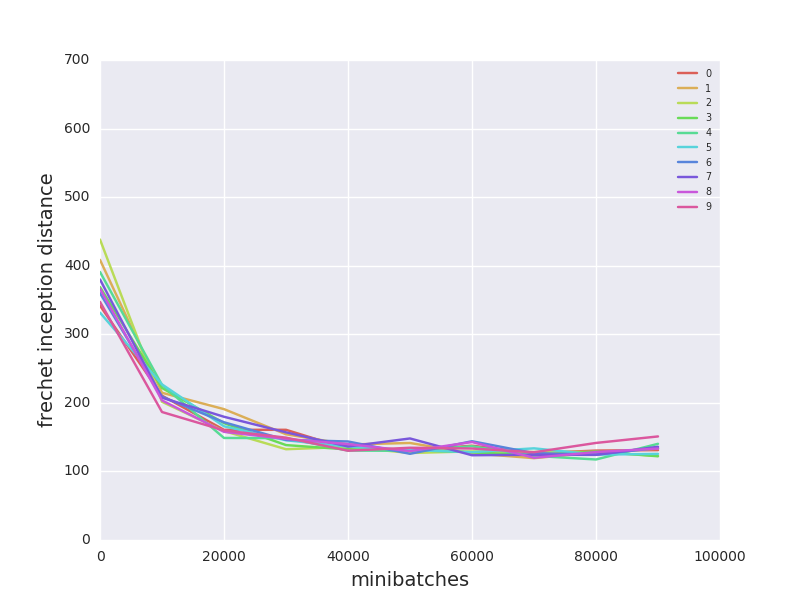}
    \label{stl:f}
  \end{subfigure}\\[\appverticalspace]
  \begin{subfigure}[t]{\appsubfigurewidth}
    \centering
    \includegraphics[width=\textwidth]{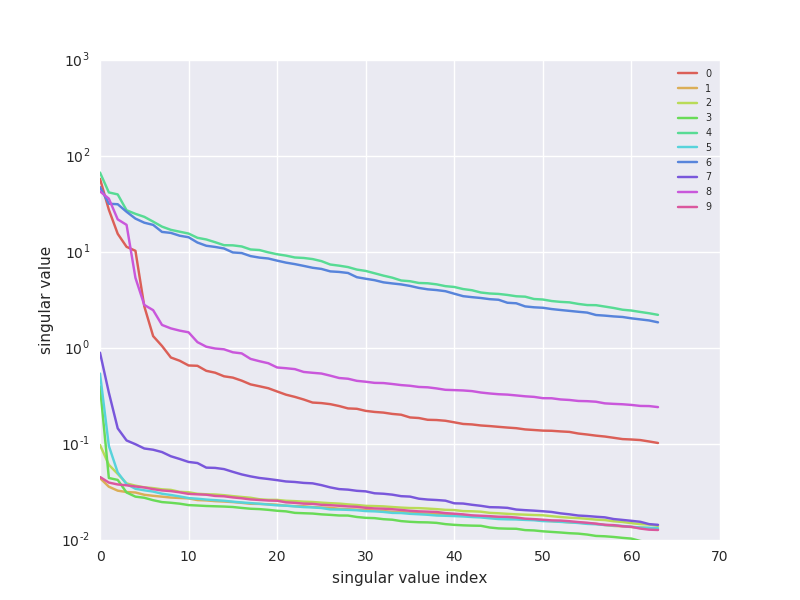}
    \label{stl:e}
  \end{subfigure}
  \begin{subfigure}[t]{\appsubfigurewidth}
    \centering
    \includegraphics[width=\textwidth]{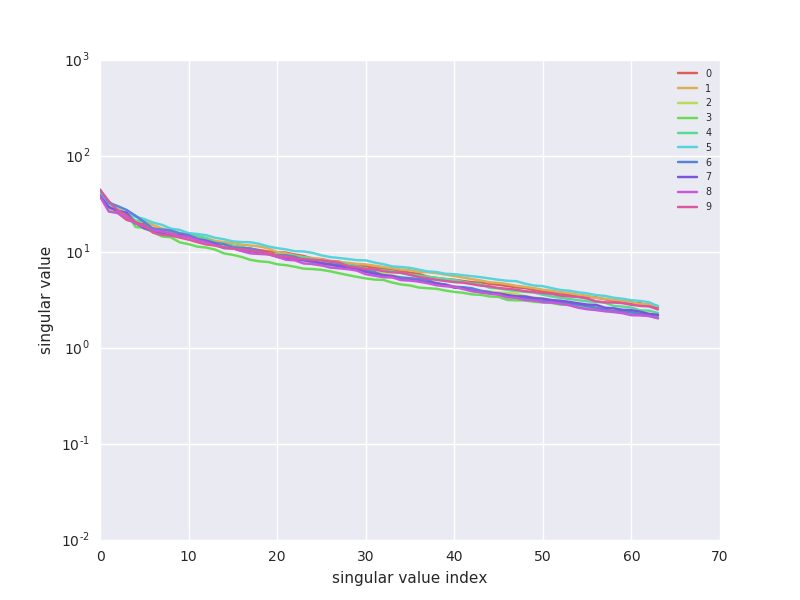}
    \label{stl:f}
  \end{subfigure}
  \vspace{-0.8cm}
  \caption{
  STL10 Experimental Results. Left and right columns correspond to 10 runs without and with Jacobian Clamping, respectively. Within each column, each run has a unique color.
  }
  \label{fig:stl}
\end{center}
\end{figure*}

%
%
%
\end{document}